\documentclass[conference]{IEEEtran}
\usepackage{times}

\usepackage[numbers]{natbib}
\usepackage{multicol}
\usepackage{amsmath}
\usepackage{booktabs}
\usepackage{graphicx} 
\usepackage{array}
\usepackage[bookmarks=true]{hyperref}
\usepackage[table]{xcolor}
\usepackage{tikz}
\usetikzlibrary{positioning,arrows.meta,calc,fit}

\usepackage{algorithm}
\usepackage{algpseudocode}

\usepackage{booktabs}
\usepackage{tabularx}
\usepackage{array}
\newcolumntype{Y}{>{\raggedright\arraybackslash}X}

\usepackage[most]{tcolorbox}
\usepackage{xcolor}

\definecolor{lightgraybox}{RGB}{245,245,245}

\newtcolorbox{definitionbox}{
    colback=lightgraybox,
    colframe=black!60,
    boxrule=0.5pt,
    arc=3pt,
    left=6pt,
    right=6pt,
    top=6pt,
    bottom=6pt,
    fonttitle=\bfseries,
}

\usepackage{xcolor}
\newlength{\mbarwd}
\newlength{\mbarht}

\newcommand{\mbar}[2]{%
  \makebox[\mbarwd][l]{\textcolor{#2}{\rule[-0.15ex]{#1\mbarwd}{\mbarht}}}%
  \hspace{2pt}%
}
\definecolor{Ransalu}{RGB}{228, 26, 28}   %
\definecolor{Xin}{RGB}{0, 0, 255}  %
\definecolor{Eren}{RGB}{77, 175, 74}   %
\definecolor{Riana}{RGB}{152, 78, 163}  %

\newlength{\barmax}

\begin{document}

\title{Physical Agentic AI: An Architecture for Orchestrating a Robot Crew with LLMs}

\author{
\IEEEauthorblockN{
  Xinyuan Liu,
  Eren Sadikoglu,
  Riana Chatterjee,
  Ransalu Senanayake
}
\IEEEauthorblockA{
  School of Computing and Augmented Intelligence, Arizona State University, AZ, USA \\
  Email: \{xinyua11, esadikog, rchatt18, rsenana1\}@asu.edu
}
}

\maketitle

\begin{abstract}
Agentic AI frameworks provide a flexible interface for interpreting open-ended task goals and decomposing them into multi-step plans. While providing the model with richer information about embodiment-specific capabilities, physical preconditions, and cross-robot coordination requirements improves theoretical grounding, we find that it does not eliminate infeasible, mistimed, or unsafe physical actions. To isolate the mechanisms of planner knowledge and execution authority, we demonstrate that physical robot crews require an explicit architectural interface between semantic planning and execution, where every planned action is verified against robot capabilities, system state, and workflow constraints before actuation. This paper introduces \textit{Physical Agentic AI, an architectural framework} for skill-grounded robot agent orchestration, in which each robot exposes a typed library of executable skills while a foundation model planner decomposes a task into phases and assigns each phase to a robot--skill pair. The framework is realized through a Robot Orchestration layer that exposes the current skill library, robot state, named locations, and workflow constraints to a non-actuating Mission Planner, while a Robot Orchestrator validates and executes one verified skill at a time. We evaluate the framework in simulation on a drone–UGV search-and-dispatch mission and on real robots in a humanoid–quadruped object transportation task. By varying planner knowledge and runtime enforcement independently, we find that retrieval substantially improves skill grounding but leaves invalid dispatch rates above 20\%. In contrast, the deterministic Robot Orchestrator reduces false dispatch rate to 0\% and blocks all injected faults across both testbeds—a result further supported by a held-plan ablation. These findings show that while retrieval substantially improves grounding, reliable physical execution requires constraints enforced outside the language model. \url{https://github.com/Liuuuxy/physical-agentic-ai}

\end{abstract}

\IEEEpeerreviewmaketitle

\section{Introduction}

Large language models (LLMs) are increasingly being used to connect natural language task specifications to robot behavior. Prior systems have used LLMs to sequence robot skills, generate executable policy code, reason with feedback, and scale robot data collection \cite{ahn2022can,huang2022inner,liang2023code,10.5555/3618408.3618748,zitkovich2023rt,ahn2024autort}. These developments herald a future in which people need only voice a goal, while robots autonomously discern its meaning, break it into purposeful steps, and carry it through to completion.

A related development in software systems is the rise of agentic AI: LLM-based agents that decompose tasks, call tools, pass messages, and coordinate multi-step workflows \cite{taparia2026learning}. Although these abstractions effectively organize complex software workflows, they are insufficient for physical robot crews, where tool calls trigger state-dependent, time-consuming, and potentially irreversible actions. As a result, robot-agent systems require an explicit architecture that bridges semantic task reasoning and physically grounded execution.

While robotics has produced several influential system architectures, including ATLANTIS \cite{gat1992integrating}, AuRA \cite{arkin1997aura}, the three-layer architecture \cite{gat1998three}, HeRO~\cite{agha2021nebula}, and ROS \cite{quigley2009ros}, the integration of foundation models introduces a novel architectural vulnerability: conflating planner knowledge with runtime safety. We argue that for heterogeneous robot teams executing autonomous workflows, isolating deliberative language reasoning from deterministic execution authority is a central research problem. Coordinating multiple embodied agents requires principled interfaces between language reasoning, robot capabilities, execution state, memory, and workflow synchronization, concerns that are not collectively addressed by existing software agentic AI abstractions.

This paper studies these requirements and the knowledge-versus-enforcement gap in the context of heterogeneous robot crews. We introduce \textit{Physical Agentic AI}, a framework for adapting agentic-AI-style orchestration to physical robots by grounding language-level plans in embodiment-specific skill libraries and workflow contracts. 
\begin{definitionbox}
\textbf{Definition (Physical Agentic AI).}
Physical Agentic AI is an agentic AI framework in which foundation model agents plan and coordinate physical robots by calling only verified robot skills, while using robot state, memory, workflow constraints, and execution checks to ensure that each planned step is physically executable.
\end{definitionbox}
As the central mechanism of this architecture, rather than allowing an LLM to generate unconstrained actions or low-level robot commands directly from language, the foundation model acts as a high-level planner over retrieved, pre-verified skills; deciding what may touch the world belongs exclusively to the deterministic Robot Orchestrator. Each robot in the crew is represented by a library of implemented skills with specified parameters, preconditions, costs, and synchronization requirements. Given a user request, the planner operates strictly over this library: it decomposes the task into phases, assigns each phase to a robot and skill, and composes a workflow-grounded plan that can be validated and executed by the robot crew.

A further reason prompt-side grounding cannot suffice is temporal: some execution-critical facts do not exist at planning time. A rover's destination may only become available after another robot completes perception; no retrieved context can validate a value that has not yet been produced. Such facts must be represented as state-bound values and resolved by the runtime at the moment of dispatch.

\begin{figure*}[h]
\centering
\includegraphics[width=\linewidth]{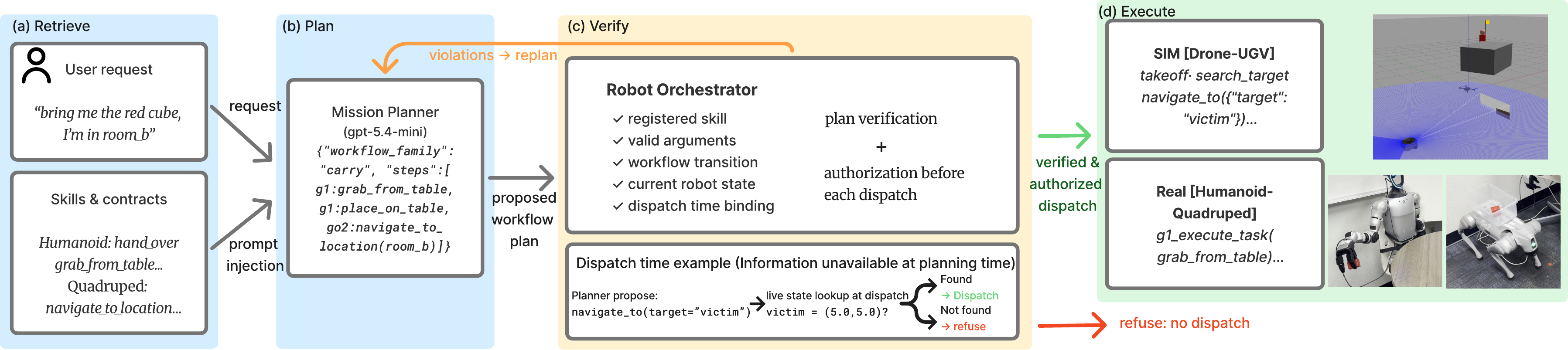}
\caption{The Physical Agentic AI pipeline. (a) Retrieval injects typed skills, locations, and workflow contracts into the planning context; (b) a non-actuating Mission Planner generates a structured plan; (c) the Robot Orchestrator validates the plan, permits one feedback replan, and re-authorizes each dispatch, rejecting invalid steps before actuation; and (d) authorized skills execute on heterogeneous platforms: a drone–UGV Gazebo team with dispatch-time state binding, and a physical humanoid–quadruped crew.}
\label{fig:architecture}
\end{figure*}

The contribution of this paper is threefold:
    
\begin{enumerate}
\item A controlled dissociation study across two heterogeneous crews---four
conditions varying only retrieved context and enforcement, plus a held-plan
ablation---showing that grounding is a retrieval problem while dispatch
safety is an enforcement problem that prompt-side knowledge does not solve.
\item An empirical characterization of \emph{retrieval-induced
substitution} and runtime binding failures: better-informed planners can shift from detectable hallucination to plausible wrong actions, motivating request-containment and state-interface checks.
\item Physical Agentic AI, the enabling architecture: typed skill
libraries, workflow contracts compiled from one declarative specification
into both the planner's prompt and the runtime checker, and per-dispatch
authorization by a deterministic orchestrator with sole actuation
authority, validated from simulation to real hardware.
\end{enumerate}

\section{Background}

Physical Agentic AI builds on four related lines of work: architectures used in robot control, software agent orchestration, foundation model robot planning, and multi-robot coordination. 

\subsection{Robot Architectures}
\label{sec:archite}
Robot software architectures have long been organized around the amount of deliberation between sensing and actuation. Three-layer architectures---a deliberative planner, a sequencing layer, and a behavioral controller---became the
dominant design by balancing long-term planning with reactive execution
\cite{gat1992integrating,gat1998three}. Over time, the sequencing layer evolved into a task executive responsible for action dispatch, precondition checking,
failure handling, and recovery through frameworks such as SMACH, FlexBE, behavior trees, and task-and-motion planning
\cite{Bohren2010TheSH,colledanchise2018behavior,garrett2021integrated,kaelbling2011hierarchical,schillinger2016flexbe}.
ROS provides the middleware foundation on which most of these systems are built
\cite{quigley2009ros,macenski2022robot}.

Our Robot Orchestrator extends this lineage rather than replacing it. Classical task executives assume that planners produce actions drawn from a closed, predefined vocabulary. Grounding is therefore a binding problem: mapping an action such as \texttt{pick(red\_cup)} to an implemented robot skill. While such an action may still fail because of sensing, reachability, or hardware constraints, it is guaranteed to refer to a capability the robot possesses.

Foundation models remove this assumption by generating plans in free-form natural language. A step such as ``Carefully organize the workspace'' may be syntactically valid yet correspond to no executable robot capability. Worse, as we demonstrate in this paper, an LLM provided with a strict capability vocabulary may still confidently dispatch unrequested or physically invalid skills. The
Orchestrator therefore introduces an explicit capability-resolution stage before conventional execution. Each proposed action must first resolve to a registered
robot capability before state, resource, and precondition checks are applied. Capability membership is established first; executability is evaluated second. We extend this model to heterogeneous multi-robot teams in Section~\ref{sec:skill_planning}.

\subsection{Software Agent Orchestration}

Artificial intelligence has evolved from systems that merely respond to prompts into ones that can reason, make decisions, and execute multi-step actions with limited human intervention. This shift has led to the emergence of Agentic AI, designed to autonomously plan, adapt, and deploy systems \cite{acharya2025agentic,li2024survey}. Software-oriented frameworks such as CrewAI, LangChain-style tool orchestration, and Model Context Protocol (MCP) tool interfaces provide practical mechanisms for connecting LLMs to external tools and shared context \cite{ayyagari2025mcp,crewai_sequential_process,hou2026mcp}. These frameworks rely on sequential process graphs to define roles, tasks, and dependencies. However, software orchestration typically assumes that tool calls are reversible, sandboxed, or easily retried; physical robot actions are not. A robot-oriented orchestration layer therefore requires explicit state tracking, physical skill availability, and execution constraints rather than only a list of callable APIs \cite{ahn2022can,kaelbling2011hierarchical}.

\subsection{Foundation Models for Robot Planning}

Foundation models have been used to interpret high-level human requests, sequence robot skills, generate robot code, and reason over feedback \cite{ahn2022can,huang2022inner,liang2023code,10.5555/3618408.3618748,zitkovich2023rt,ahn2024autort}. SayCan established a central paradigm for this setting: language model reasoning is constrained by a library of pretrained robot skills and their affordance/value estimates, allowing semantic knowledge to be combined with embodiment-specific feasibility~\cite{ahn2022can}. This capability-grounding perspective has since been extended from individual robots to heterogeneous robot teams.

SMART-LLM uses an LLM to decompose high-level instructions, form robot coalitions, and allocate subtasks across a heterogeneous team~\cite{kannan2024smart}. RoCo similarly uses LLM-based multi-robot dialogue and subtask planning, but additionally incorporates environment feedback such as collision checking into iterative plan refinement~\cite{mandi2024roco}. COHERENT extends this pattern to heterogeneous platforms through a proposal--execution--feedback--adjustment loop in which a centralized task assigner distributes subtasks to robot executors and updates the plan from execution feedback~\cite{liu2025coherent}. EMOS further emphasizes embodiment-aware reasoning by constructing machine-readable descriptions of robot capabilities from URDF and kinematic information, enabling heterogeneous agents to reason over physically grounded capabilities~\cite{chen2025emos}. More recently, EmboTeam combines LLM instruction parsing with PDDL planning and reactive behavior trees, using a shared blackboard to coordinate dynamically sized heterogeneous teams~\cite{zeng2026emboteam}. Collectively, these systems demonstrate that foundation models can provide useful semantic reasoning, capability grounding, task allocation, and feedback-driven coordination for multi-robot systems.

A complementary line of work addresses the boundary between language-level reasoning and physical execution. Classical robot architectures and task executives separate deliberative planning from execution-time state checking and recovery~\cite{gat1998three,colledanchise2018behavior,garrett2021integrated}. SafeGate makes this separation explicit for LLM-controlled robots by introducing deterministic pre-execution safety gates and task safety contracts that reject commands or unsafe state transitions before or during execution~\cite{obi2026pre}. RoCo and COHERENT also incorporate forms of feasibility or execution feedback, but these mechanisms primarily support plan construction or reactive correction~\cite{mandi2024roco,liu2025coherent}. Our work focuses on a related but distinct interface for heterogeneous robot crews: the LLM is non-actuating, and every proposed robot--skill transition is independently admitted by a deterministic runtime using both robot-local state and cross-robot workflow state.

The distinction is important when a mission contains constraints that span robots and evolve during execution. For example, a quadruped may be permitted to depart only after a humanoid has completed loading, or a rover destination may become available only after another robot performs a search. Such conditions are not properties of an isolated skill or a completed plan; they are properties of the joint workflow at the moment a transition is executed. We therefore represent these dependencies as executable cross-robot workflow contracts and give the Robot Orchestrator sole actuation authority. In this formulation, retrieval improves what the planner can propose, whereas runtime authorization determines which proposed transitions are admitted to the physical system.

\subsection{Multi-Robot Coordination and Physical Grounding}

Multi-robot systems inherently require robust task allocation, communication, temporal coordination, and collision avoidance \cite{fernando2022graphical, fernando2022coco}. Classical planning and task-and-motion-planning (TAMP) methods provide structured ways to reason about discrete goals and continuous feasibility \cite{lavalle2006planning,kaelbling2011hierarchical}. Formal methods can rigidly encode temporal mission constraints and safety specifications \cite{kress2009temporal}. These methods offer absolute guarantees and interpretability, but they typically rely on carefully specified models that are difficult to connect directly to open-ended natural language requests.
Foundation models address a complementary need: interpreting
underspecified human instructions. Frameworks for general-purpose agent orchestration such as CrewAI, AutoGen, and MCP-based tool servers have already implemented role assignment, tool and skill registration, and communication among agents
\cite{crewai_sequential_process,wu2023autogen,ayyagari2025mcp,hou2026mcp}, which is analogous to how we treat role assignments and skill libraries for software agents. What these frameworks do not support is physical state or irreversibility: an unsuccessful tool invocation can simply be retried or ignored. Task-and-motion planning, by contrast, takes explicit account of physical preconditions and effects, but requires predefined symbolic domains and offers no support for reasoning with an open-ended language model \cite{kaelbling2011hierarchical,garrett2021integrated}. Our
architecture combines elements of both, applying the roles and skill libraries of software agent orchestration together with the state and precondition reasoning of TAMP to planning with language models. The resulting system uses agentic orchestration for task delegation while preserving the classical robot-planning constraints needed for physical execution \cite{crewai_sequential_process,kaelbling2011hierarchical,kress2009temporal}.

\section{The Architecture}
\label{sec:skill_planning}

To isolate the mechanisms of planner knowledge and execution authority, our architecture separates semantic reasoning from physical execution. A non-actuating \emph{Mission Planner} interprets the user request and proposes a workflow over the capabilities of the robot crew, while a deterministic \emph{Robot Orchestrator} (RO) decides whether each proposed action may actually be dispatched. This functional separation creates a single execution boundary: foundation models may propose actions, but only the RO has authority to actuate a robot.

The distinction is important in physical systems for two reasons. First, a plausible action generated by an LLM is not necessarily grounded in an implemented robot capability or valid in the current physical state. Second, some execution-critical information may not exist when the mission is planned. For example, a rover's destination may only become available after a drone completes a perception task. Such information must be resolved from runtime state rather than inferred or reconstructed by the planner.

By instantiating this separation through robot skill libraries, a foundation model Mission Planner, executable workflow contracts, and a deterministic Robot Orchestrator, the architecture explicitly disentangles two mechanisms that are often conflated in foundation model robotics: \emph{retrieval}, which provides the planner with grounded knowledge of the current system, and \emph{enforcement}, which determines whether a proposed action is authorized at execution time.

\subsection{Design Principles}

Rather than introducing another foundation model planning algorithm, we focus on the architectural boundary connecting semantic planning to physical execution. The design follows four principles:

\paragraph{Separation of reasoning and execution}
Foundation models excel at interpreting natural language and decomposing tasks~\cite{ahn2022can,huang2022inner} but they cannot reliably self-enforce physical constraints. Planning and physical execution are therefore assigned to different system components with no overlap in authority.

\paragraph{Skill-grounded planning}
The planner reasons only over implemented robot skills rather than generating arbitrary actions or low-level control commands. 
Each robot exposes a bounded library of executable skills governed by typed parameters, preconditions, and expected effects.

\paragraph{Contract-mediated coordination}
Multi-robot workflows are represented as reusable workflow contracts that encode ordering constraints, synchronization points, and resource dependencies. These contracts capture coordination logic independently of the language model.

\paragraph{Execution-time verification}
Every selected skill is validated against the workflow contract and current
robot state immediately before actuation, ensuring only safe, grounded actions cross the reality gap.

\subsection{Architectural Components}

Figure~\ref{fig:architecture} illustrates the overall architecture, which consists of four principal components.

\paragraph{Robot Skill Libraries}
A planner can only be stopped from inventing actions if the system maintains a strict inventory of real capabilities to check against. Each robot $r$ exposes a typed library $K_r$ of executable skills. A skill defines an implemented robot capability along with its arguments, preconditions, expected effects, execution cost, and synchronization requirements. For example, a humanoid may expose \texttt{pick\_place(object, location)}, while a quadruped exposes \texttt{navigate(location)}. 

\paragraph{Mission Planner}
The Mission Planner is a non-actuating foundation model agent responsible for understanding the user request, decomposing it into task \emph{phases}, selecting robot--skill pairs, and producing an executable workflow~\cite{lewis2020retrieval,li2025llm}. Because it never communicates directly with robot controllers, it cannot physically dispatch unsafe actions.

Given a user request and system state, the planner produces a plan:
\begin{equation}
\pi = \big[(\phi_1, r_1, \kappa_1),\ \ldots,\ (\phi_T, r_T, \kappa_T)\big],
\end{equation}
\noindent where step $k$ pairs a phase $\phi_k$ with a robot $r_k$ and one of its skills $\kappa_k$. A plan is \emph{admissible} only if every step names a real skill with valid arguments, the assigned robot is able and available, and the operational ordering holds (e.g., a carrier must be loaded before it departs). When several admissible plans exist, the system prefers the simplest, with the fewest robots and handoffs. Our prototype keeps this concrete by having the planner choose among a few enumerated workflows rather than searching a large space; the architecture is independent of that choice, and a stronger planner would occupy the same role behind the same checks.

\paragraph{Workflow Contracts}

A workflow contract encodes the coordination constraints that must hold during execution. 

\begin{definitionbox}
\textbf{Definition (Executable Workflow Contract).}
An executable workflow contract is a declarative specification that
mediates between semantic mission planning and physical execution. It
specifies the participating robot roles and admissible skills, guarded
ordering constraints between them, argument-validity rules, and
state-bound values resolved at dispatch time.
\end{definitionbox}

A contract specifies synchronization points, temporal dependencies, resource
constraints, completion conditions, and execution invariants~\cite{kress2009temporal,kaelbling2011hierarchical}. Crucially, while the planner selects and instantiates a contract, enforcement belongs exclusively to the orchestrator's runtime.

\paragraph{Robot Orchestrator (RO)}

In software agentic AI, \emph{orchestration} typically means routing a tool call and passing results between agents. In Physical Agentic AI it additionally requires execution authority. The Robot Orchestrator (RO) provides this authority. It is a deterministic procedural runtime, not a foundation model. It receives the structured plan $\pi$ and the instantiated workflow contract, maintains the execution state, and controls access to robot-specific execution interfaces. 

Immediately before dispatching a skill, it verifies that:
\begin{enumerate}
\item the requested skill exists on the selected robot; \item the skill arguments are grounded and valid, and satisfy \emph{request containment}: user-specified entities or destinations cannot be replaced with different registered values merely to make the plan executable; \item the requested step is the next authorized transition in the workflow contract; \item the current robot state satisfies the skill and contract preconditions;\item any runtime-bound values required by the step are available and valid. \end{enumerate}

Only after these checks succeed does the RO forward the action to the hardware execution interface. This structural distinction guarantees that providing constraints to a foundation model (retrieval) is fully decoupled from trusting the model to obey them (enforcement). The RO treats the generated plan strictly as a proposal, not as a command stream.

\subsection{Execution Protocol}

Given a user request, execution proceeds through a sequence of verified stages:

\begin{enumerate}
\item Retrieve the current skill libraries, named locations, and workflow
      contracts, injecting them into the planning context.
\item The Mission Planner interprets the request and generates a structured
      workflow consisting of an instantiated contract and an ordered list of robot--skill--argument steps.
\item The RO parses the plan and validates the entire workflow globally (skill/location grounding and contract ordering). If validation fails, one replan is permitted before the mission is rejected.
\item Execute approved skills sequentially; each dispatch is re-authorized against the
    next pending step of the validated workflow and consumed on authorization, so no step is silently retried. Robot state updates from the structured feedback of each blocking call.
\item The workflow terminates when all steps complete. A runtime contract
      violation is refused at the gate before actuation; a failed executed
      step halts the mission and reports it, rather than attempting automatic recovery.
\end{enumerate}

\subsection{Implementation}
We instantiate this architecture with CrewAI~\cite{crewai_sequential_process} as the base agent framework, modified to enforce the planning--execution separation. The Mission
Planner's prompt is augmented with the retrieved context (skills, locations, contracts) rendered from the exact same declarative specifications that the runtime enforces. This ensures the rules the model reads and the rules the orchestrator checks cannot drift. The Robot Orchestrator is implemented as procedural code with sole actuation
authority. Robot adapters map these authorized skills to embodiment-specific commands over ROS2 and Unitree SDK2 communication. Because the high-level tool signatures remain consistent, the planner and contracts remain entirely unchanged across both simulation and real robot deployments.

\section{Experiments}

\label{sec:exp}

We evaluate Physical Agentic AI across two complementary testbeds with deliberately different roles. The \emph{simulation} testbed (Section~\ref{sub:air_ground}; an Iris quadcopter and a TurtleBot3) serves as our scalable quantitative evaluation: it compares RO against the LLM-only and skill-list baselines, scales to many trials, and injects controlled faults to rigorously measure safety-gate behavior. The \emph{hardware} testbed (Section~\ref{sec:hw_crew}; a Unitree G1 humanoid and a Unitree Go2 quadruped) provides physical validation, demonstrating that the same skill-grounded planning architecture transfers to real, irreversible execution on heterogeneous physical embodiments.

\subsection{Air--Ground Orchestration in Simulation}
\label{sub:air_ground}

\begin{figure}[t]
\centering
\includegraphics[trim=160 37 240 147, clip, width=0.78\linewidth]{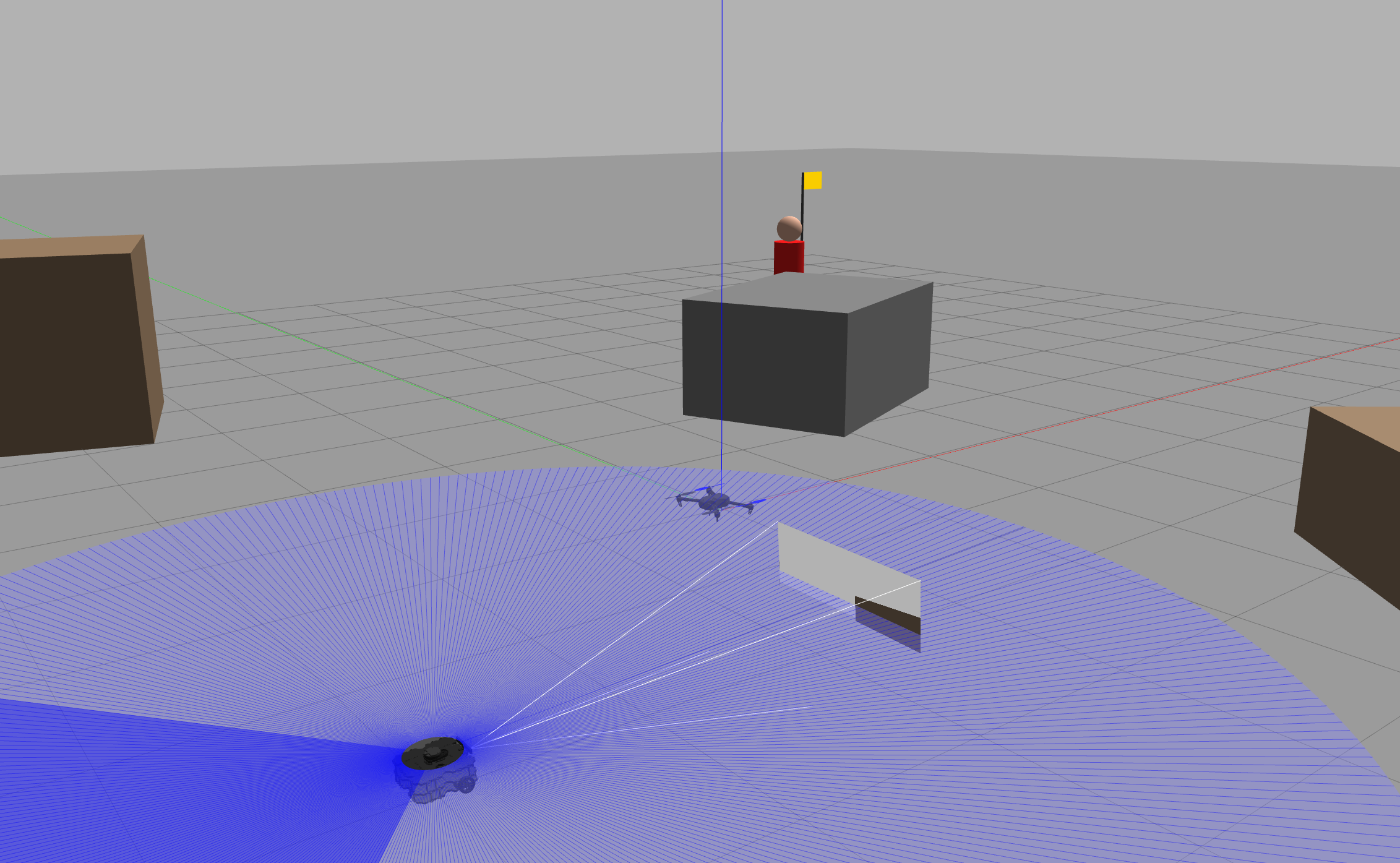}
\caption{Gazebo \textsc{Search-and-Dispatch} testbed: the Iris drone
localizes the victim (flag) while the workflow contract holds the
TurtleBot3 (foreground, lidar shown) staged until a validated detection
is published.}
\label{fig:sim_env}
\end{figure}

\begin{figure*}[t]
\centering
\includegraphics[width=0.96\textwidth]{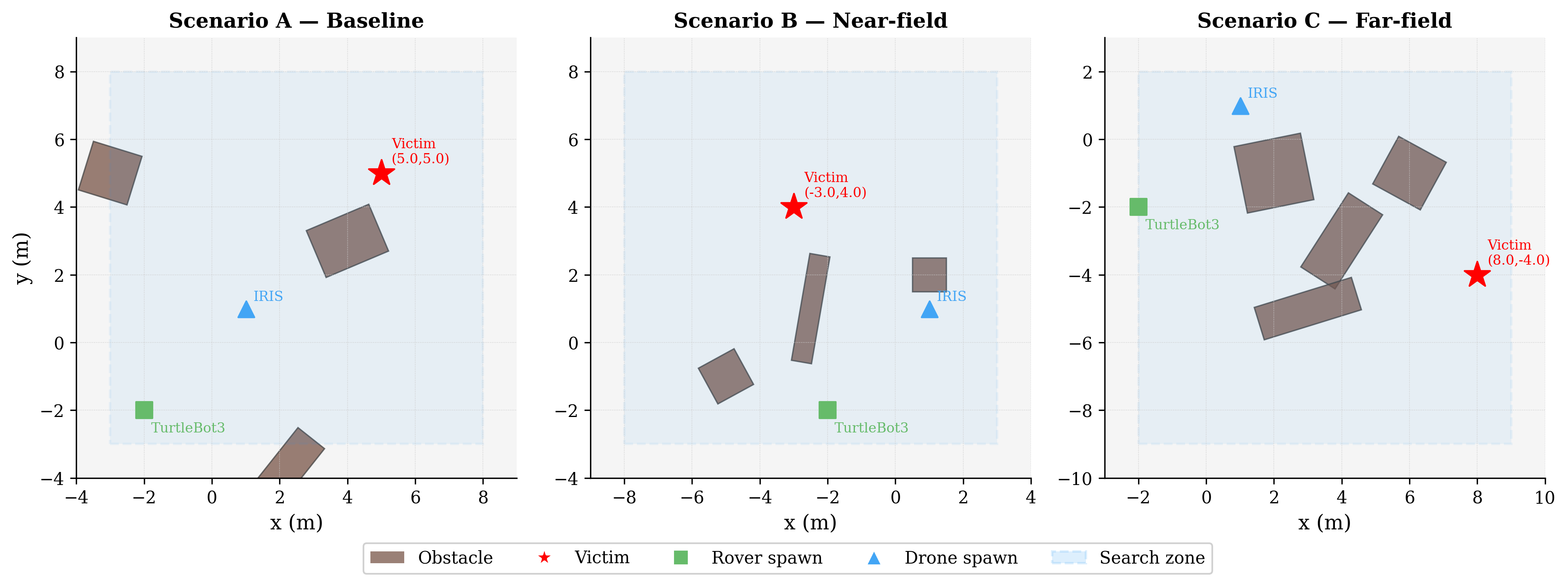}
\caption{The three air--ground \textsc{Search-and-Dispatch} evaluation
scenarios. Each panel shows the search zone (shaded), obstacles, the Iris
drone and TurtleBot3 spawn poses, and the victim location: \textbf{A
(Baseline)} places the victim at $(5,5)$, \textbf{B (Near-field)} at
$(-3,4)$, and \textbf{C (Far-field)} at $(8,-4)$, varying the distance and
obstacle clutter between the rover spawn and the delivery goal. All
configurations in Table~\ref{tab:sim-comparison} are evaluated on these
three maps.}
\label{fig:sim_scenarios}
\end{figure*}

The air--ground testbed is the controlled test of the proposed orchestration architecture: an Iris quadcopter that localizes a victim from the air and a
TurtleBot3 rover that drives to the published position, in Gazebo under ROS2
with PX4 SITL and MAVROS (Fig.~\ref{fig:sim_env};
Fig.~\ref{fig:sim_scenarios} shows the three scenario maps). Beyond
establishing the knowledge-versus-enforcement dissociation, it probes a failure mechanism unique to multi-robot sensing: the decisive fact---\emph{the victim's position}---does not exist at planning time, so no amount of prompt-side knowledge can validate the rover's goal prior to execution. The planner, contracts, orchestrator, and evaluation harness are shared with the humanoid--quadruped system of Section~\ref{sec:hw_crew}; only the skill libraries, workflow contracts, and robot adapters differ.

\subsubsection{Experimental Setup}

\paragraph{Platforms and Skills}
Each embodiment exposes a typed skill library that defines the planner's
entire action space. The Iris exposes \texttt{takeoff(altitude)}, \texttt{search\_target(bounds, altitude)}, \texttt{get\_coordinates()}, and \texttt{land()}. The TurtleBot3 exposes \texttt{navigate\_to(target $|$ x, y)} and \texttt{stop()}. Coordinates are strictly bounded to the declared operational area ($x,y\in[-6,10]$\,m); requests outside this region fail grounding before execution.

\paragraph{Dynamic Goal Binding} Because the rover's goal is produced by the drone's search at run time, the plan binds it \emph{symbolically} (e.g., \texttt{navigate\_to(target="victim")}). The RO resolves this binding against the state interface at dispatch. If no finite fix has been published, the step is refused (\texttt{no\_target\_fix}). 
The planner reasons over the published
fix and never processes raw perception data.

\paragraph{Workflow Contracts}
Four workflow contracts
(\textsc{Search-and-Dispatch}, \textsc{Search-only}, \textsc{Dispatch-only},
\textsc{Abort}) define admissible compositions.
\textsc{Search-and-Dispatch}, for example, structurally orders takeoff before search and localization before dispatch. The contract prose shown to the planner is
rendered directly from the same declarative specifications that the deterministic gate enforces.

\subsubsection{Evaluation Protocol}
\label{sec:eval_design}
Prior evaluations cannot tell whether safety gains come from what the
planner \emph{knows} or what the runtime \emph{enforces}, because both
change together. We separate them with four conditions:
\begin{itemize}
    \item \emph{LLM-only}: No skill registry is provided to the planner.
    \item \emph{Skill-list}: The skill registry and operational area are retrieved and provided. 
    \item \emph{RO-prompt}: Skill and workflow constraints are provided in context, but are not enforced at execution.
    \item \emph{RO}: The same planning context is used with dispatch-time enforcement and one feedback-driven replan. \end{itemize}

All conditions use the same foundation model (\texttt{gpt-5.4-mini}, temperature
0.1), output schema, parser, and dispatch adapter. We evaluate on 20 scenarios: 12 nominal (three per workflow family) and 8 fault injections. The fault injections include four robot-unavailable states (two per robot, checked only at dispatch), two out-of-bounds destinations, and two perception faults (\textit{nan\_transmission}: the victim is detected but the published fix is NaN-corrupted; \textit{missed\_detection}: the sweep completes without a detection). Perception faults are invisible
to static plan checking---they surface only on the state interface at the moment of rover's dispatch.

\subsubsection{Results}
\paragraph{Result 1: Grounding and Safety Are Dissociable}
Table~\ref{tab:sim-comparison} establishes the core dissociation. Retrieval
repairs grounding: injecting the registry raises skill grounding from 51\%
to 96\% ($p{=}9.7{\times}10^{-7}$). Without it, nearly half of the
LLM-only planner's actions are entirely hallucinated. However, once knowledge
saturates (Skill-list, RO-prompt, and RO are statistically identical on
every knowledge metric), the non-enforcing arms still dispatch 23\% of grounding- or state-faulted steps and block no fault scenario (0\% recall). Only enforcement changes the outcome: RO eliminates false
dispatch entirely (0\%, $p{=}5.0{\times}10^{-4}$ vs.\ RO-prompt) and correctly blocks all eight injected fault scenarios with no false blocks (100\% recall and precision). Notably, RO-prompt and RO share the exact same prompt; they differ only in whether a flawed step is granted physical actuation authority. Because every condition was executed live, these false dispatches are executed robot actions rather than rejected plan strings; Section~\ref{sec:gazebo-ex} reports what the eight fault missions did physically under the unenforced arm.

\paragraph{Result 2: Closing the Structured-Channel Bypass}
The perception faults show why enforcement must sit at the dispatch boundary. When a corrupted coordinate (\textit{nan\_transmission}) is flagged, an unenforced LLM will attempt to recover an approximate position from the surrounding text and dangerously dispatch the rover anyway. The model replaces a value it cannot ground with a plausible concrete one that passes static name-level checks. By resolving the goal against the state interface, RO closes this bypass: an absent or non-finite fix triggers a hard \texttt{no\_target\_fix} refusal, blocking the perception faults in all runs (Table~\ref{tab:sim-comparison}).

\begin{table*}[t]
\centering
\caption{Air--ground \textsc{Search-and-Dispatch} evaluation across 20 scenarios (12 nominal, 8 fault-injection). Every mission in every condition was executed live in Gazebo against PX4 SITL and the ROS2 robot services. All conditions share the model, schema, parser, and dispatch interface. Brackets denote 95\% Wilson confidence intervals; Fisher's exact tests compare RO against each baseline. Darker bars denote RO on a $0$--$100\%$ scale.}
\label{tab:sim-comparison}
\small
\setlength{\tabcolsep}{3pt}
\begin{tabular}{lllll}
\toprule
Metric & LLM-only & Skill-list & RO-prompt & RO\\
\midrule
\multicolumn{5}{l}{\textit{Planner behavior (what was proposed)}}\\
Workflow-selection acc.\ (\%)
& \mbar{1.00}{gray!45}100 [84, 100] & \mbar{1.00}{gray!45}100 [84, 100]
& \mbar{1.00}{gray!45}100 [84, 100] & \mbar{1.00}{black!80}100 [84, 100]\\

Skill grounding (\%)
& \mbar{0.52}{gray!45}51 [37, 65] & \mbar{0.96}{gray!45}96 [86, 99]
& \mbar{0.96}{gray!45}96 [86, 99] & \mbar{0.96}{black!80}\textbf{96 [86, 99]}\\

Plan executability (\%)
& \mbar{0.25}{gray!45}25 [11, 47] & \mbar{0.90}{gray!45}90 [70, 97]
& \mbar{0.90}{gray!45}90 [70, 97] & \mbar{0.90}{black!80}\textbf{90 [70, 97]}$^{\ddagger}$\\

Contract violation (\%) $\downarrow$
& \mbar{0.75}{gray!45}70 [48, 85] & \mbar{0.004}{gray!45}0 [0, 16]
& \mbar{0.004}{gray!45}0 [0, 16] & \mbar{0.004}{black!80}0 [0, 16]\\
\midrule
\multicolumn{5}{l}{\textit{Execution outcomes (what reached the robots)}}\\
False dispatch (\%) $\downarrow$
& \mbar{0.54}{gray!45}53 [39, 67] & \mbar{0.25}{gray!45}23 [14, 37]
& \mbar{0.23}{gray!45}23 [14, 37] & \mbar{0.004}{black!80}\textbf{0 [0, 8]}\\

Safety-gate recall (\%)
& \mbar{0.004}{gray!45}0 [0, 32] & \mbar{0.004}{gray!45}0 [0, 32]
& \mbar{0.004}{gray!45}0 [0, 32] & \mbar{1.00}{black!80}\textbf{100 [68, 100]}\\
\midrule
Safety-gate precision (\%)
& \mbar{0.004}{gray!45}0 [0, 0] & \mbar{0.004}{gray!45}0 [0, 0]
& \mbar{0.004}{gray!45}0 [0, 0] & \mbar{1.00}{black!80}\textbf{100 [68, 100]}\\

Planning latency (s)
& $1.4{\pm}0.4$ & $1.2{\pm}0.5$ & $2.2{\pm}0.5$ & $3.0{\pm}1.1$\\
\bottomrule
\end{tabular}

\vspace{0.4em}
\begin{minipage}{0.98\textwidth}
\footnotesize
$^{\dagger}$Safety-gate precision is not meaningful for the three baselines because they never produce positive safety-gate decisions; it is therefore reported descriptively rather than used for statistical comparison.
$^{\ddagger}$Plan executability is bounded at 90\% by design: in the two
out-of-bounds fault scenarios every planner correctly copies the requested
(invalid) destination, so those plans are ungrounded in all conditions.
Replans recover none of the fault scenarios because all eight faults are
unrecoverable by construction; RO refuses and reports rather than
silently substituting a feasible mission.
\end{minipage}
\end{table*}

\begin{figure}[t]
\centering
\includegraphics[width=\linewidth]{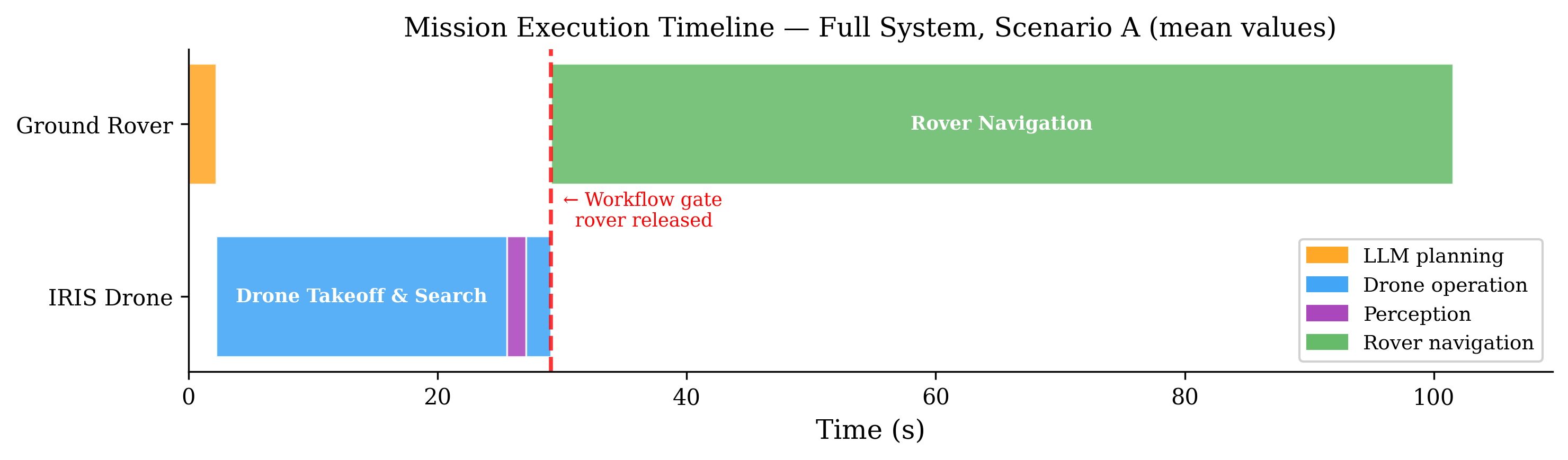}
\caption{Execution timeline (full system, Scenario~A, means). The workflow
gate (dashed) holds the rover idle until the drone's detection is published,
then releases navigation; cf.\ the hardware timeline in Fig.~\ref{fig:timeline}.}
\label{fig:sim_timeline}
\end{figure}

\subsubsection{Gazebo Execution Validation}
\label{sec:gazebo-ex}
Every mission in Table~\ref{tab:sim-comparison} was executed live in Gazebo against PX4 SITL and the ROS2 robot services, so each reported dispatch reflects an execution outcome rather than a static plan check. Under the full system, the execution timeline in Fig~\ref{fig:sim_timeline} shows the rover held until the drone publishes a detection; the planner generated the expected takeoff–search–dispatch sequence, and the rover reached the published goals with a mean error of $0.465$\,m ($0.44$–$0.49$\,m), comparable to $0.494$\,m for a scripted non-LLM run. With a NaN-transmission fault, the corrupted coordinate failed dispatch-time validation and the rover remained at its spawn position.

The unenforced RO-prompt arm shows what this gate prevents. Across the eight injected faults, all eight violating steps were dispatched and six produced robot motion. Robots marked unavailable were commanded anyway, while the out-of-bounds rover continued toward an invalid destination until the $300$\,s mission deadline. The remaining two produced no motion only because the underlying navigation service rejected the malformed goals. Thus, the false dispatches reported in Table I correspond to commands that crossed the orchestration boundary, not merely invalid plan strings.

\begin{figure*}[t]
\centering
\setlength{\tabcolsep}{2pt}
\renewcommand{\arraystretch}{0.7}
\begin{tabular}{cccc}
\multicolumn{4}{c}{\textbf{(a) Near requester: direct humanoid handoff}} \\[2pt]

\includegraphics[width=0.228\textwidth]{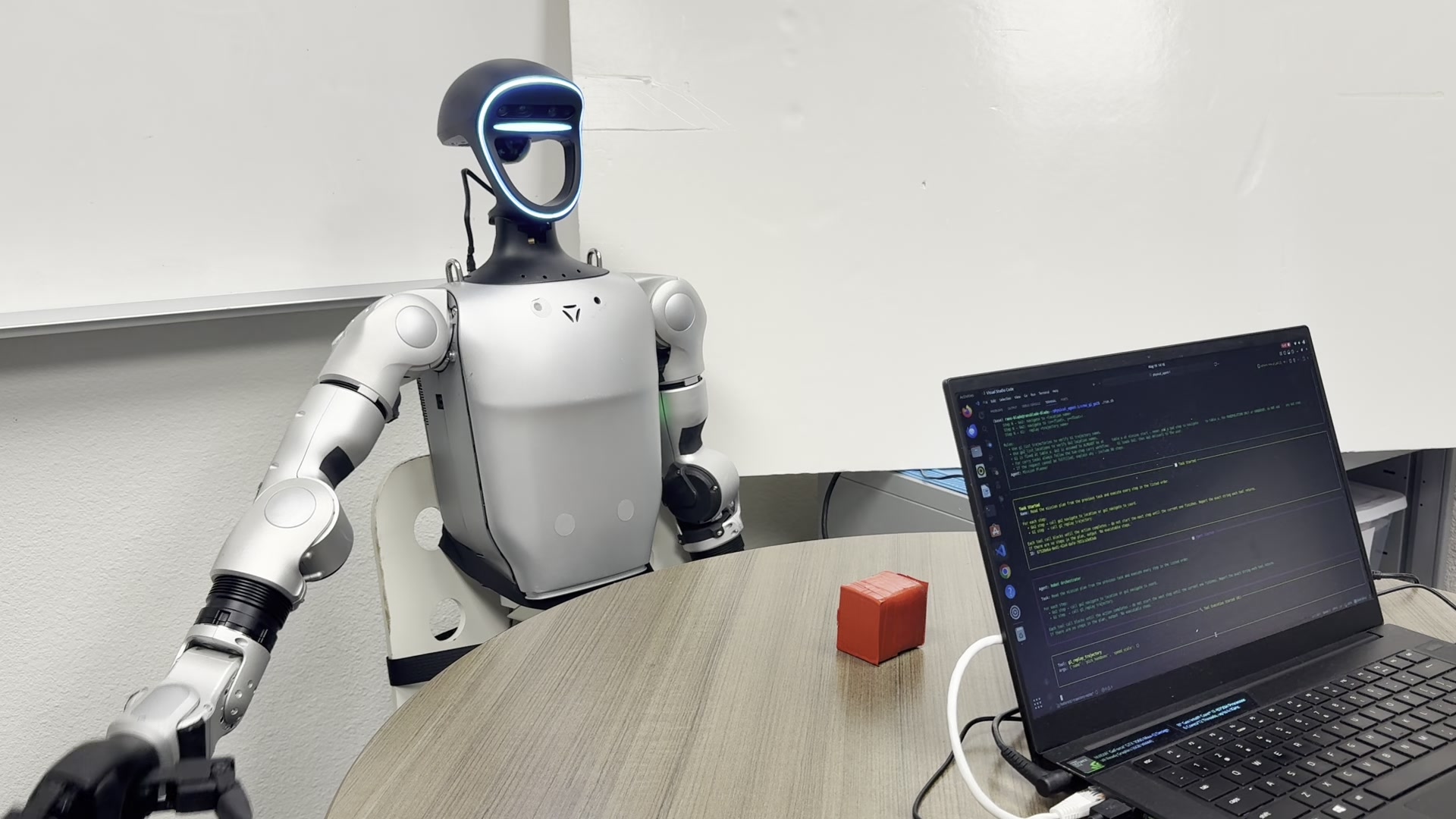} &
\includegraphics[width=0.228\textwidth]{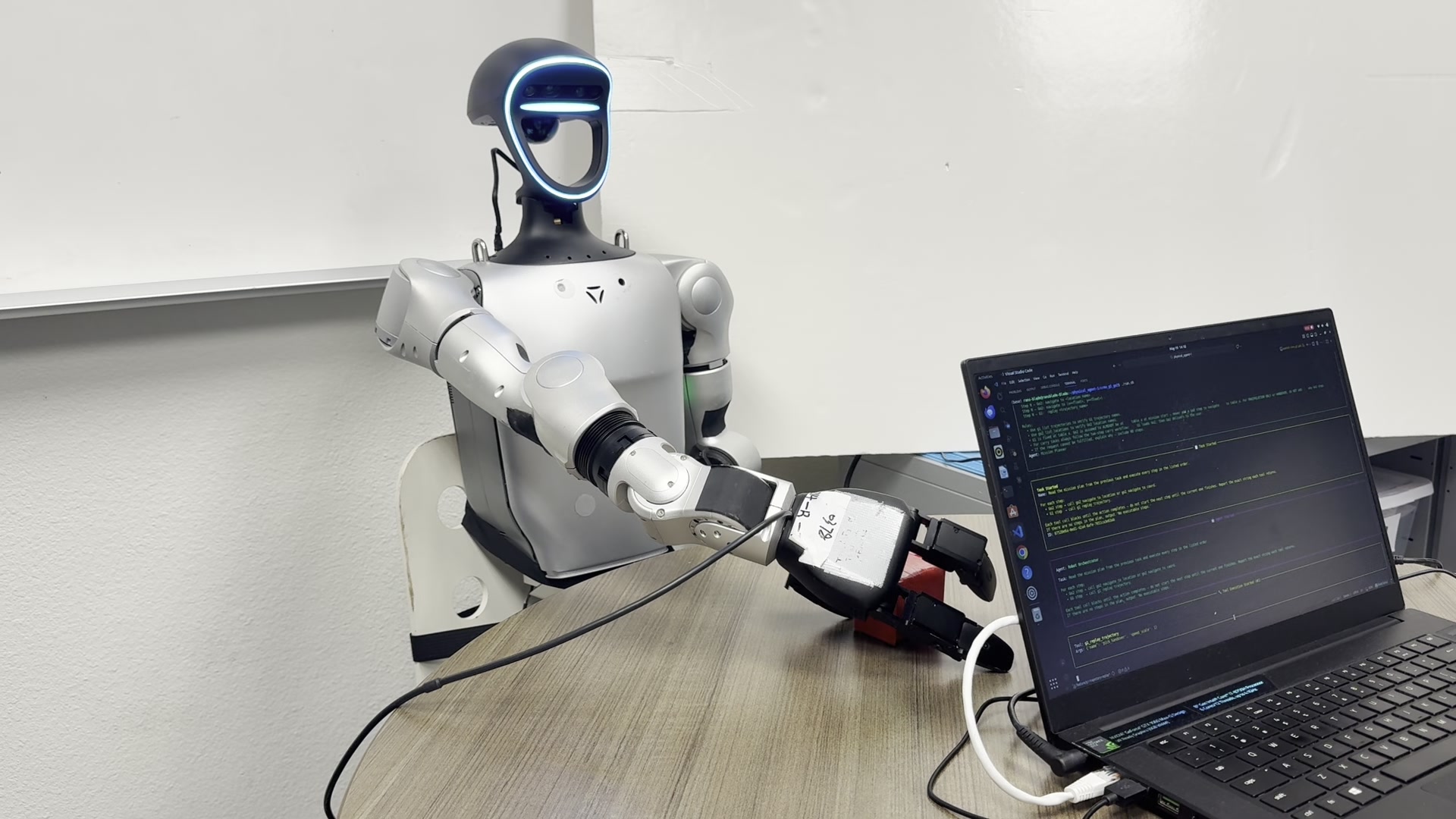} &
\includegraphics[width=0.228\textwidth]{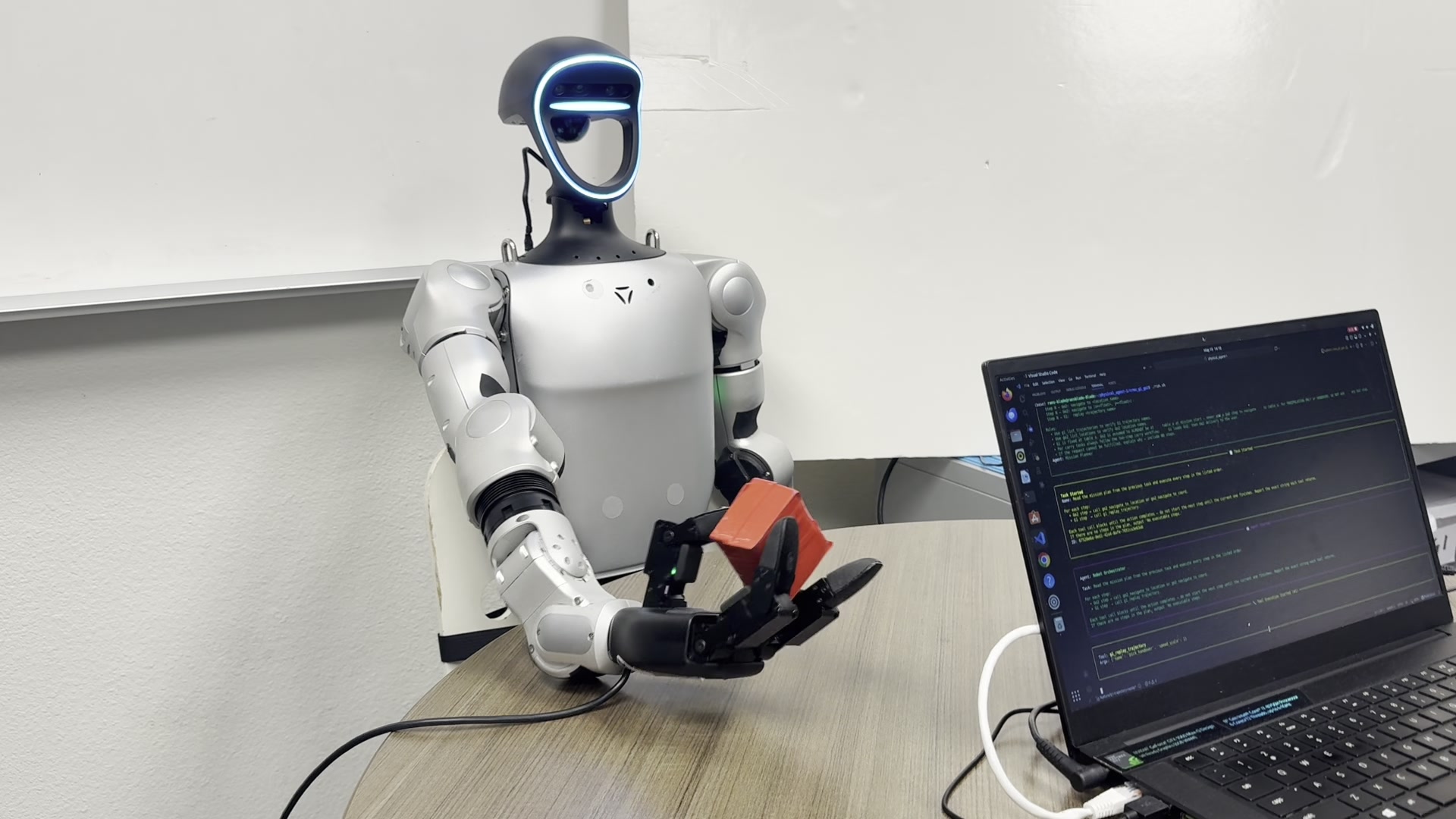} &
\includegraphics[width=0.228\textwidth]{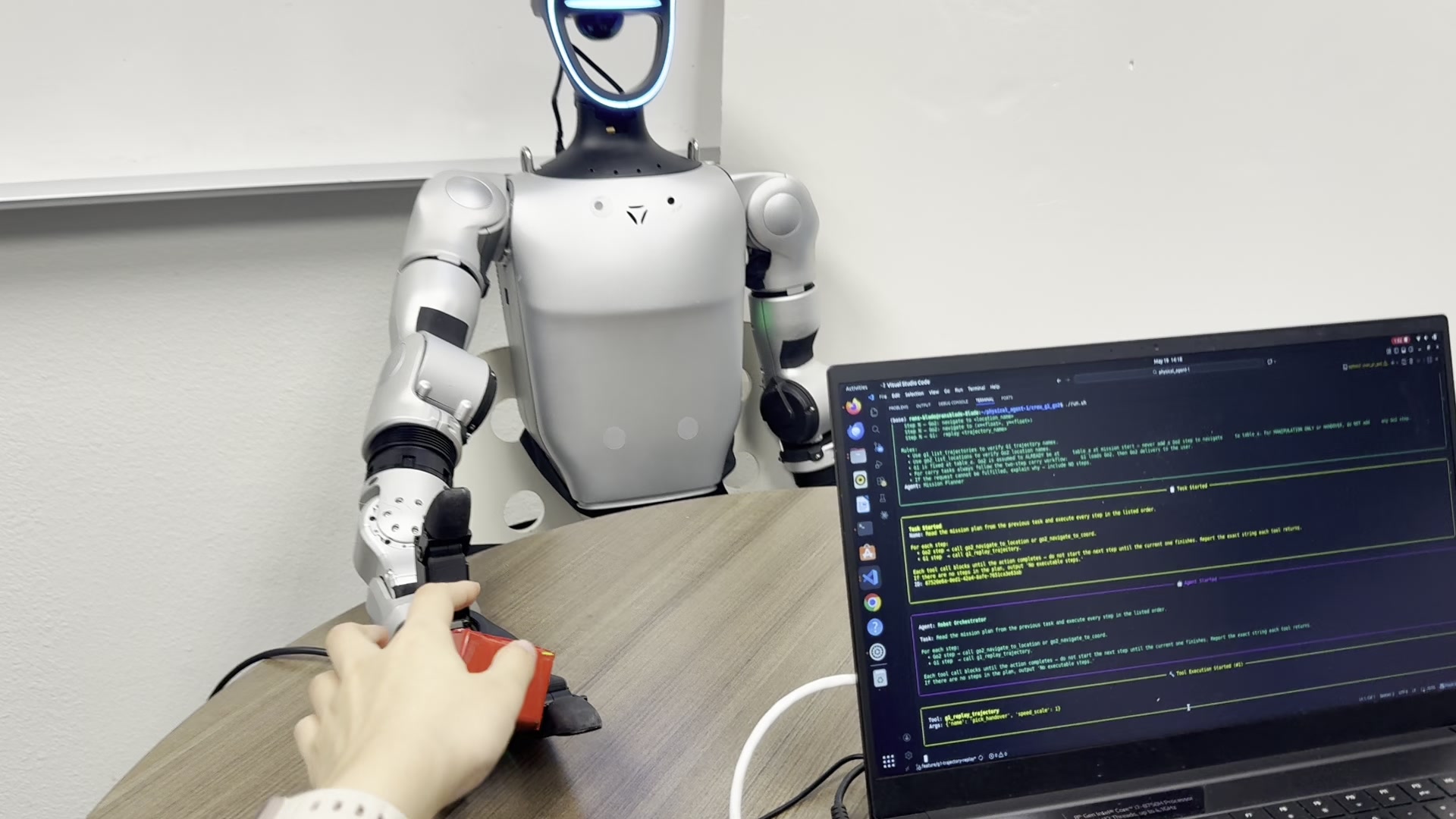} \\

{\scriptsize\shortstack{(1) Request + plan\\$\approx$0--15\,s}} &
{\scriptsize\shortstack{(2) Humanoid reaches\\$\approx$15--19\,s}} &
{\scriptsize\shortstack{(3) Grasps red cube\\$\approx$20--23\,s}} &
{\scriptsize\shortstack{(4) Direct handoff\\$\approx$29--32\,s}} \\[4pt]

\multicolumn{4}{c}{\textbf{(b) Far requester: humanoid--quadruped delivery}} \\[2pt]

\includegraphics[width=0.228\textwidth]{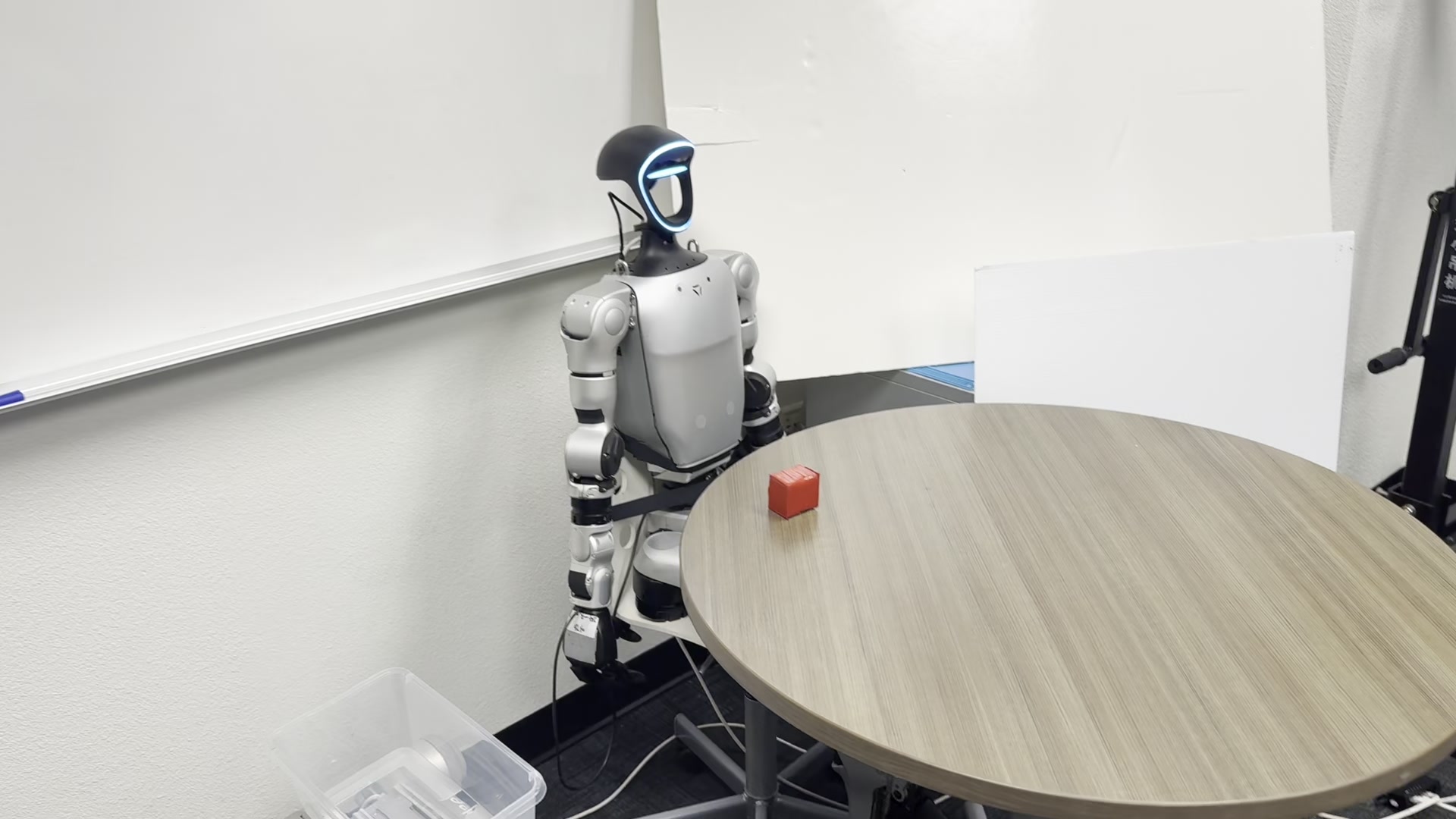} &
\includegraphics[width=0.228\textwidth]{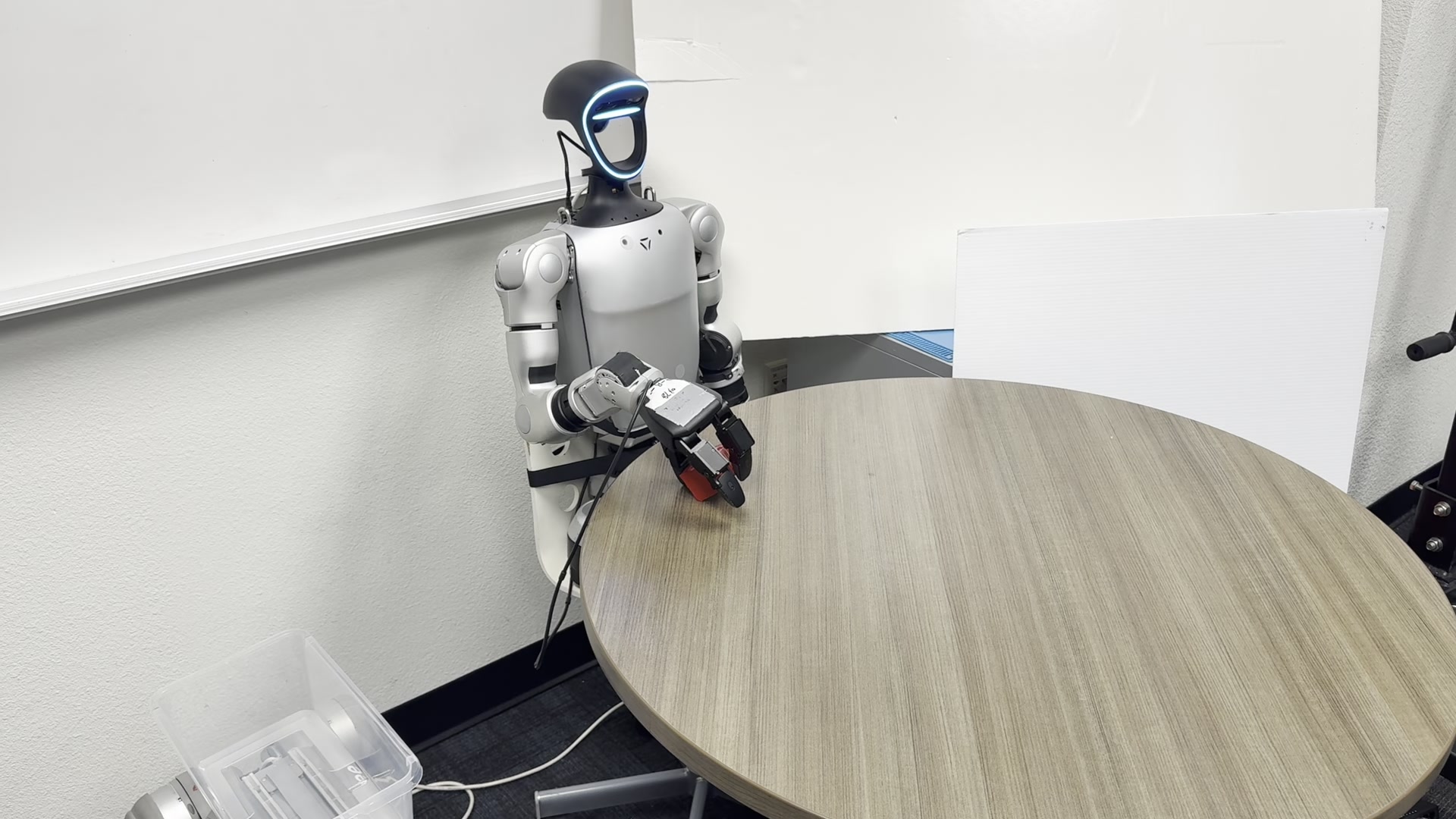} &
\includegraphics[width=0.228\textwidth]{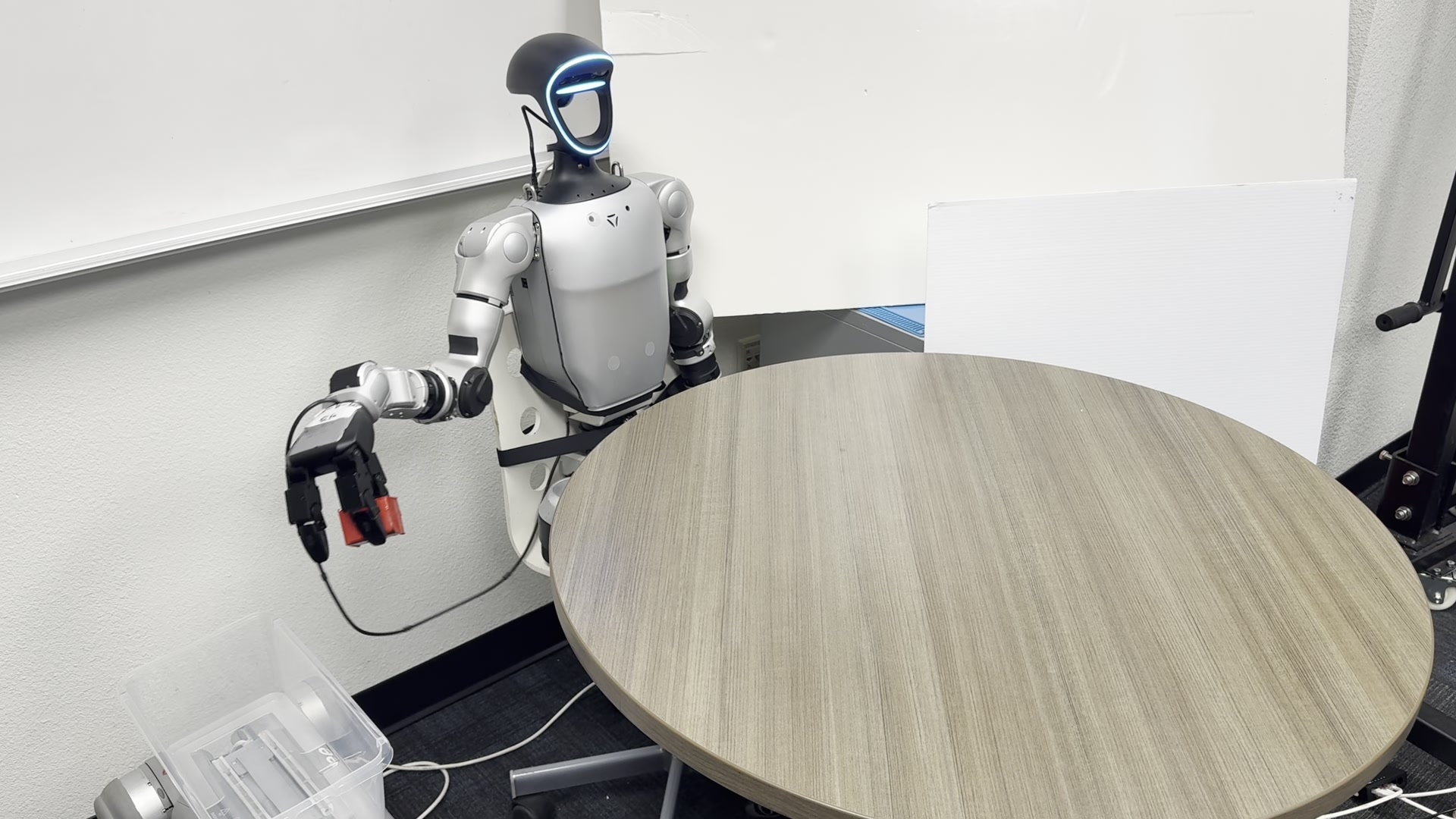} &
\includegraphics[width=0.228\textwidth]{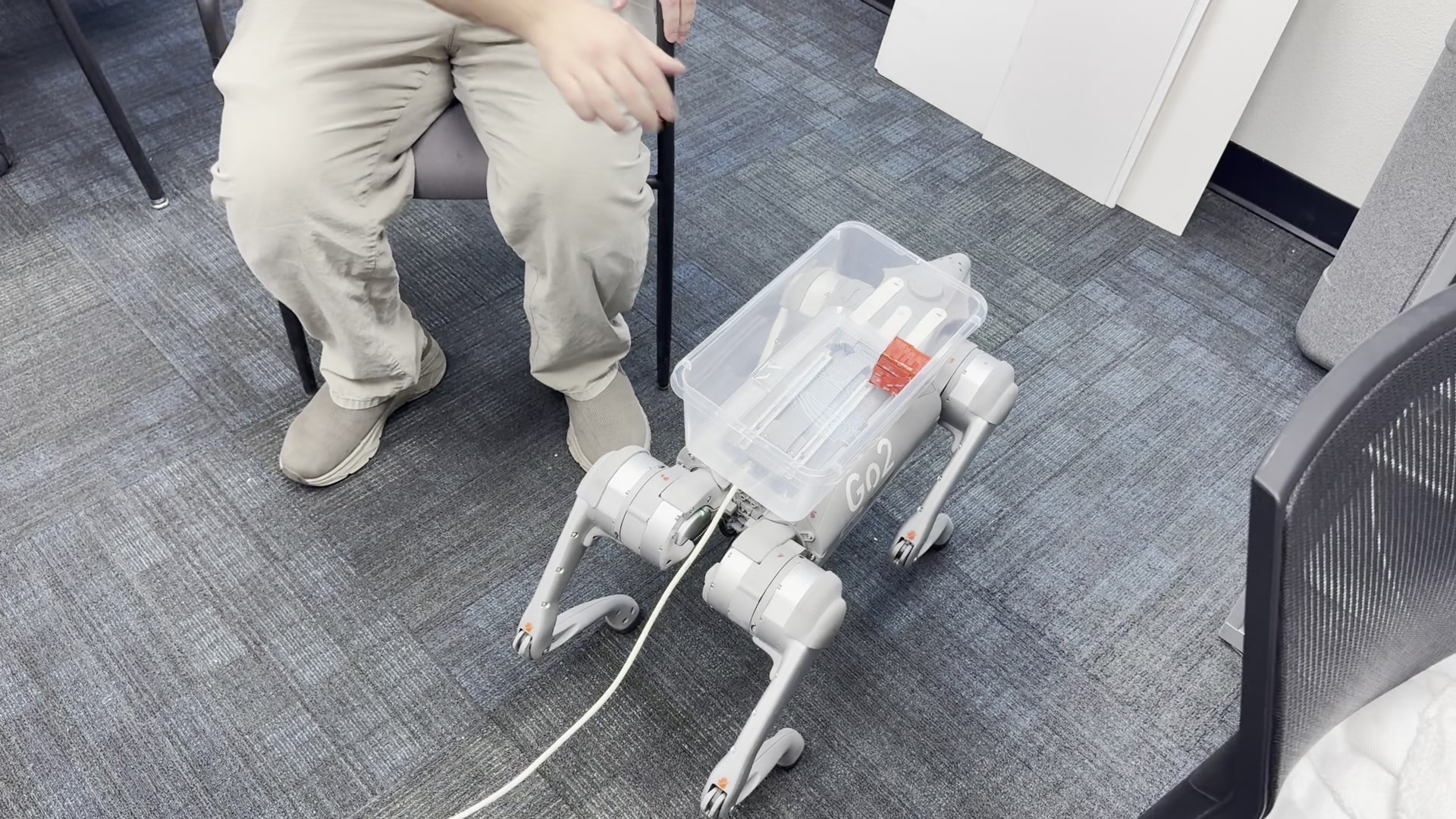} \\

{\scriptsize\shortstack{(1) Request + plan\\$\approx$0--6\,s}} &
{\scriptsize\shortstack{(2) Humanoid grasps cube\\$\approx$6--14\,s}} &
{\scriptsize\shortstack{(3) Loads cube onto Go2\\$\approx$14--27\,s}} &
{\scriptsize\shortstack{(4) Quadruped delivers\\$\approx$30--41\,s}} \\
\end{tabular}

\caption{Hardware demonstrations of distance-conditioned delivery, with representative frames extracted from the two recordings.
Top (a): when the requester is near the humanoid, the planner selects the \textsc{Handover} workflow and the humanoid (Unitree G1) hands the red cube directly to the person.
Bottom (b): when the requester is far from the humanoid, the planner selects the \textsc{Carry} workflow; the humanoid grasps the red cube and loads it onto the quadruped (Unitree Go2) payload bin, which then navigates across the room to deliver it. Approximate phase windows annotated from the clips are shown beneath each frame.}
\label{fig:hardware_demo}
\end{figure*}

\subsection{Humanoid--Quadruped Crew}
\label{sec:hw_crew}

The humanoid--quadruped testbed translates this dissociation study to physical robots.
A Unitree G1 humanoid provides tabletop manipulation and a Unitree Go2
quadruped provides room-scale transport. The quantitative study evaluates 20 scenarios using mocked adapters over the hardware stack's tools and checking code, while real-robot trials (Section~\ref{sec:hw_demos}) validate physical transfer.

\subsubsection{Platform and Skill Libraries}

The G1 is treated as a manipulation agent at a known table location. Its skill library contains replayable manipulation skills such as \texttt{pick\_handover}, which hands an object to a nearby requester, and \texttt{pick\_place}, which loads an object onto the Go2 payload tray. The Go2 is treated as a mobile transport agent that can navigate among named locations such as \texttt{table\_a} and \texttt{room\_b}. These named skills and locations define the action space available to the planner.

Four workflow contracts (\textsc{Handover}, \textsc{Carry},
\textsc{Navigation-only}, \textsc{Manipulation-only}) define the admissible
compositions, including the \textsc{Carry} rule that the quadruped may not
depart before the humanoid loads it. We report workflow-selection accuracy, skill grounding, plan executability, contract-violation attempts, false dispatches, safety-gate precision/recall, and planning latency. This metric set separates language-level task interpretation from physical executability and safety checking.

\begin{table*}[t]
\centering
\caption{Mocked humanoid--quadruped evaluation across 20 scenarios (12 nominal, 8 fault-injection). The setup uses hardware-equivalent skill interfaces with mock adapters. Brackets denote 95\% Wilson confidence intervals. Fisher's exact tests compare RO against each baseline. Darker bars denote RO on a $0$--$100\%$ scale.}
\label{tab:g1_go2_mock_eval}
\small
\setlength{\tabcolsep}{4pt}
\begin{tabular}{lllll}
\toprule
Metric & LLM-only & Skill-list & RO-prompt & RO\\
\midrule
\multicolumn{5}{l}{\textit{Planner behavior (what was proposed)}}\\
Workflow-selection acc.\ (\%)
& \mbar{1.00}{gray!45}100 [84, 100] & \mbar{1.00}{gray!45}100 [84, 100]
& \mbar{1.00}{gray!45}100 [84, 100] & \mbar{1.00}{black!80}100 [86, 100] \\
Skill grounding (\%)
& \mbar{0.09}{gray!45}9 [3, 22] & \mbar{0.91}{gray!45}91 [77, 97]
& \mbar{0.97}{gray!45}97 [85, 100] & \mbar{0.97}{black!80}\textbf{97 [85, 100]} \\
Plan executability (\%)
& \mbar{0.15}{gray!45}15 [5, 36] & \mbar{0.85}{gray!45}85 [64, 95]
& \mbar{0.95}{gray!45}95 [76, 99] & \mbar{0.95}{black!80}\textbf{95 [76, 99]} \\
Contract violation (\%) $\downarrow$
& \mbar{0.70}{gray!45}70 [48, 85] & \mbar{0.10}{gray!45}10 [3, 30]
& \mbar{0.15}{gray!45}15 [5, 36] & \mbar{0.15}{black!80}15 [5, 36] \\
\midrule
\multicolumn{5}{l}{\textit{Execution outcomes (what reached the robots)}}\\
False dispatch (\%) $\downarrow$
& \mbar{0.94}{gray!45}94 [81, 98] & \mbar{0.29}{gray!45}29 [17, 46]
& \mbar{0.26}{gray!45}26 [14, 42] & \mbar{0.004}{black!80}\textbf{0 [0, 10]}\\
Safety-gate recall (\%)
& \mbar{0.004}{gray!45}0 [0, 32] & \mbar{0.004}{gray!45}0 [0, 32]
& \mbar{0.004}{gray!45}0 [0, 32] & \mbar{1.00}{black!80}\textbf{100 [68, 100]} \\
Safety-gate precision (\%)
& \mbar{0.004}{gray!45}0 [0, 0] & \mbar{0.004}{gray!45}0 [0, 0]
& \mbar{0.004}{gray!45}0 [0, 0] & \mbar{1.00}{black!80}\textbf{100 [68, 100]}\\
\midrule
Planning latency (s)
& $1.14{\pm}0.35$ & $2.51{\pm}1.71$ & $2.60{\pm}1.44$ & $2.99{\pm}1.62$\\
\bottomrule
\end{tabular}

\vspace{0.4em}
\begin{minipage}{0.98\textwidth}
\footnotesize
$^{\dagger}$Safety-gate precision is degenerate for the two baselines because they do not issue positive safety-gate decisions in these fault cases; we therefore interpret this metric descriptively rather than as evidence of a meaningful precision comparison. Contract-violation attempts denote proposed actions or states that would violate the workflow contract; under RO these are refused by the Orchestrator rather than executed.
\end{minipage}
\end{table*}

\subsubsection{Results}
\paragraph{Result 1: The Dissociation Replicates via Held-Plan Ablation}
Table~\ref{tab:g1_go2_mock_eval} reproduces the simulation pattern on physical hardware infrastructure. Retrieval
repairs grounding (9\% to 91\%), but the non-enforcing arms still blindly dispatch 26--29\% of faulted steps. RO enforcement drops false dispatch to 0\% and raises recall to 100\% ($p{=}1.6{\times}10^{-4}$).

To definitively isolate causality, we conducted a {held-plan ablation}:
we sampled the planner once per scenario and dispatched the identical plans through both the RO-prompt and RO modes, mathematically removing planner variance. Routing the same plans through the deterministic gate moved the false dispatch rate from 25.7\% (9/35 steps) down to
0\%.

\paragraph{Result 2: Retrieval-Induced Substitution}
\label{sec:substitution}
Providing more grounding information also changes how the planner fails. For
the request \emph{``carry the red cube to the rooftop helipad,''} (an invalid location), LLM-only invents a nonexistent skill. Skill-list selects a real skill but retains the invalid destination, \texttt{place\_on\_table(``rooftop\_helipad'')}. However, with the full registry provided, the RO-prompt planner silently substitutes the destination to a valid but incorrect location (e.g., \texttt{navigate\_to\_location(``dropoff'')}).

We refer to this failure mode as
\emph{retrieval-induced substitution}: as planner context increases, detectable hallucinations evolve into plausible but incorrect substitutions that bypass standard syntactic grounding checks. The RO detects and safely rejects this via a strict request-containment rule.

\subsubsection{Hardware Validation}
\label{sec:hw_demos}
We validated the architecture on the Unitree G1 humanoid and Go2 quadruped using
the same request, ``bring me the red cube,'' while changing only the
requester's location. When the requester is near the table, the planner
selects \textsc{Handover} and the G1  delivers the cube directly. When the requester is farther away, the planner selects \textsc{Carry}: the G1 loads the cube onto the Go2, which then transports it
across the room to the requester (Fig.~\ref{fig:hardware_demo}).

\begin{figure}[t]
\centering
\footnotesize
\begin{tikzpicture}[x=0.18cm,y=0.85cm]
\draw[->,gray!70] (0,-0.15) -- (43,-0.15);
\foreach \t in {0,5,10,15,20,25,30,35,40} {
  \draw[gray!70] (\t,-0.10) -- (\t,-0.20);
  \node[below,font=\scriptsize] at (\t,-0.18) {\t};
}
\node[font=\scriptsize] at (21.5,-0.85) {time (s)};
\node[left,font=\scriptsize] at (-0.5,1.45) {Near};
\fill[gray!20]   (0,1.2)  rectangle (15,1.7);
\fill[blue!30]   (15,1.2) rectangle (29,1.7);
\fill[orange!55] (29,1.2) rectangle (32,1.7);
\fill[gray!20]   (32,1.2) rectangle (42,1.7);
\draw (0,1.2) rectangle (42,1.7);
\node[font=\scriptsize] at (7.5,1.45) {plan / idle};
\node[font=\scriptsize] at (22,1.45) {G1 \texttt{pick\_handover}};
\node[left,font=\scriptsize] at (-0.5,0.55) {Far};
\fill[gray!20]   (0,0.3)  rectangle (6,0.8);
\fill[blue!30]   (6,0.3)  rectangle (27,0.8);
\fill[gray!12]   (27,0.3) rectangle (30,0.8);
\fill[green!45]  (30,0.3) rectangle (40,0.8);
\fill[orange!55] (40,0.3) rectangle (41,0.8);
\draw (0,0.3) rectangle (41,0.8);
\node[font=\scriptsize] at (3,0.55) {idle};
\node[font=\scriptsize] at (16.5,0.55) {G1 grasp + \texttt{pick\_place}};
\node[font=\scriptsize] at (35,0.55) {Go2 navigate};
\end{tikzpicture}
\caption{Execution timelines for the two hardware demonstrations.
\textsc{Handover} uses only the G1, whereas \textsc{Carry} serializes G1
manipulation and Go2 transport under the load-before-depart constraint.
Timings are annotated from the recorded demonstrations.}
\label{fig:timeline}
\end{figure}
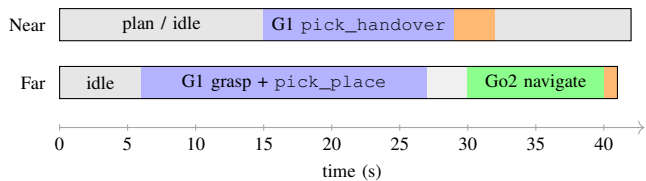

In both trials, every physical action passes through the same dispatch gate
evaluated in the mocked study. In the \textsc{Carry} workflow, the
cross-robot load-before-depart constraint prevents Go2 navigation from being
authorized until the G1 completes loading. The resulting execution sequences
are shown in Fig.~\ref{fig:timeline}: \textsc{Handover} requires only the
humanoid, whereas \textsc{Carry} serializes humanoid manipulation and
quadruped transport.

These demonstrations provide qualitative evidence that the same
planning--enforcement interface transfers to heterogeneous physical robots
and adapts execution to the physical context. Because only two hardware
trials are performed, we reserve statistical claims for the mocked
evaluation.

\subsubsection{Discussion}

The experiments support a division of labor between foundation model reasoning and robot skill execution. The foundation model interprets the request and selects a workflow; RO constrains that choice using retrieved skills, named locations, robot state, and workflow contracts; the Robot Orchestrator executes only validated tool calls. The mocked evaluation measures this decomposition at scale, while the hardware trials show that the same abstraction can drive a physical humanoid--quadruped crew.

The pattern in Table~\ref{tab:g1_go2_mock_eval} is directional rather than
incremental: exposing a skill list mainly repairs grounding, whereas workflow contracts and checked execution are what eliminate false dispatches and recover the fault cases. For hardware, the phase timeline is preferable to a latency table because there are only two trials and the main point is the change in workflow structure, not a statistically meaningful timing distribution. Future evaluations should add repeated physical trials before making hardware-level statistical claims, and separate planning errors from manipulation or navigation failures using the same staged success decomposition used in the mocked tests. The gate guarantees only the constraints represented by its capability, argument, workflow, state, and runtime-binding checks; physical failures within an authorized skill, such as a failed grasp or navigation failure, remain the responsibility of downstream execution and recovery mechanisms.

\section{Conclusion}

We asked which mechanism actually makes LLM-operated robot crews safe, and the answer is only partly what one might expect. Retrieval is necessary: without the skill registry, grounding drops sharply. But retrieval alone is not enough. Once the planner is informed, grounding improves, yet unsafe dispatches still persist at 23--29\%. The decisive difference comes from per-dispatch enforcement by a deterministic runtime, which removes false dispatches entirely and achieves 100\% fault recall on both testbeds. In live Gazebo execution, all eight injected faults crossed the orchestration boundary without enforcement and six produced robot motion, whereas enforcement refused all eight before the violating action executed. In practice, this means that language models can help with plan quality, but they should not hold execution authority. That authority belongs in the runtime, including custody over channels whose values only exist at execution time. Two questions remain open: how substitution errors vary across foundation models, and how far the declarative contract layer can scale as skill libraries grow beyond hand-authored registries.

\bibliographystyle{plainnat}
\bibliography{references}

\end{document}